\documentclass[fleqn,10pt]{wlscirep}
\usepackage[utf8]{inputenc}
\usepackage[T1]{fontenc}
\usepackage{graphicx}
\usepackage{amssymb}

\usepackage{lineno}
\usepackage{float}
\usepackage{multirow}
\usepackage{makecell}
\usepackage{mathtools}
\usepackage{amssymb}
\usepackage{amsthm}
\usepackage{bm}
\usepackage{mathrsfs}
\usepackage{amscd}
\usepackage{microtype}
\usepackage{xcolor}
\usepackage{graphicx}
\usepackage{subfigure}
\usepackage{amsmath}
\usepackage{caption}
\usepackage{booktabs} 
\usepackage[normalem]{ulem}

\title{GNN-based physics solver for time-independent PDEs}

\author[1]{Rini Jasmine Gladstone}
\author[2]{Helia Rahmani}
\author[2]{Vishvas Suryakumar}
\author[1]{Hadi Meidani}
\author[3,*]{Marta D'Elia}
\author[2,*]{Ahmad Zareei}
\affil[1]{Civil and Environmental Engineering, University of Illinois Urbana-Champaign, IL,USA}
\affil[2]{Meta Reality Labs, Redmond, WA, USA}
\affil[3]{Pasteur Labs, Brooklyn, NY, USA}

\affil[*]{marta.delia@simulation.science, azar@meta.com}

\begin{abstract}
Physics-based deep learning frameworks have shown to be effective in accurately modeling the dynamics of complex physical systems with generalization capability across problem inputs. However, time-independent problems pose the challenge of requiring long-range exchange of information across the computational domain for obtaining accurate predictions. In the context of graph neural networks (GNNs), this calls for deeper networks, which, in turn, may compromise or slow down the training process. In this work, we present two GNN architectures to overcome this challenge - the Edge Augmented GNN and the Multi-GNN. We show that both these networks perform significantly better (by a factor of 1.5 to 2) than baseline methods when applied to time-independent solid mechanics problems. Furthermore, the proposed architectures generalize well to unseen domains, boundary conditions, and materials. Here, the treatment of variable domains is facilitated by a novel coordinate transformation that enables rotation and translation invariance.
By broadening the range of problems that neural operators based on graph neural networks can tackle, this paper provides the groundwork for their application to complex scientific and industrial settings.
\end{abstract}
\begin{document}

\flushbottom
\maketitle
%
%
\thispagestyle{empty}


\section{Introduction}
\label{intro}

Numerical models for solving partial differential equations (PDEs) are crucial to scientific modeling in a broad range of fields, including physics, biology, material science, and finance. While various techniques, such as finite element methods, finite difference methods, finite volume methods, etc., have been developed as high-fidelity solvers, accelerating and reducing their computational cost remains a challenge. Additionally, when the governing PDEs are unknown, predicting a physical system using classical techniques is not possible even when observations are available. Recent developments in deep learning have enabled faster and accurate algorithms to evaluate the response of a physical system by exploiting the governing equations, observational data, or both \cite{thuerey2021pbdl,karniadakis2021physics}. 

Neural networks trained by adding the governing PDEs and boundary conditions to the loss function are a prominent example of using deep-learning-based surrogate models to learn solutions of physical system  \cite{lagaris1998artificial, RAISSI2019686, sirignano2018dgm, yu2018deep, khoo2021solving}. Here, the neural network is used as an approximate representation of the solution of the PDE. While these surrogates are useful to find the specific solution of a PDE with a given set of parameters, a slight modification of such parameters, boundary conditions, or domain geometry requires re-training, making them less attractive in settings such as optimization, design, and uncertainty quantification.

Another class of neural network-based solvers is convolutional neural networks (CNNs) which use snapshots of observed data over a discretized domain to predict the physical solution. While such data-driven methods do not require a priori knowledge of the governing PDE, they are often limited to the specific domain discretization and cannot be easily generalized to other domain geometries. CNN-based methods include PDE-inspired architectures \cite{ruthotto2020deep}, autoregressive dense encoder-decoder networks \cite{geneva2020modeling}, and  symbolic multi-layer neural networks \cite{long2019pde}.

In another class of data-driven, deep-learning-based surrogates, neural networks are used to learn a discretization that is then used in classical solvers\cite{bar2019learning, kochkov2021machine, zhuang2021learned, han2018solving}. Specifically, the neural network learns a field used for interpolation at coarse scales. Although these methods have been shown to improve the accuracy and further accelerate traditional solvers, it should be noted that they are still limited by the initial discretization

With the purpose of obtaining resolution invariance, a class of neural network-based PDE solvers focuses on continuous graph representation of the computational domain \cite{li2020neural, li2020multipole,you2022nonlocal, iakovlev2020learning, belbute2020combining}. In such representations, the continuous nature of the network update guarantees the invariance of the architecture with respect to the resolution of the data and the enhanced interaction among nodes, typical of graph-based networks, improves the accuracy of the architecture for complex physical systems. Related works have shown that the location of the graph nodes can be optimized to better learn the solution at different levels of precision \cite{alet2019graph}. Furthermore, some of these methods and their extensions \cite{li2020fourier,li2022geoFNO}, focus on learning a continuous mapping between input and output of a PDE and are referred to as {\it neural operators}. These and other neural operators, such as DeepONets \cite{lu2021learning}, have shown significant success in accurately predicting the solution of several PDEs and they generalize relatively well with respect to the PDE input parameters. We mention that one advantage of using graph-based representations is that the  system’s dynamics can be recovered with sparse representation \cite{poli2019graph, li2020fourier, iakovlev2020learning}. 

Another class of networks that exploits graph-based representation is given by MeshGraphNet and its extensions. Firstly introduced in\cite{DBLP:journals/corr/abs-2010-03409} for time-dependent problems and then extended to a multi-scale setting in \cite{{multiscale}}, these mesh-based GNNs encode in a connected graph the mesh information and the corresponding physical parameters such as loading and boundary conditions and model parameters. In addition to mesh-based approaches, it is worth mentioning that particle-based methods and their graph representation have also been been successfully used in physics simulations \cite{pmlr-v119-sanchez-gonzalez20a}. Here, the physical system is described using particles in a reduced-order-model manner; the particles are then identified as nodes of a graph and a message passing neural network is used to learn and compute the dynamics of the particles. Both these graph-based approaches can accurately approximate physics simulation and generalize well to different resolutions and boundary conditions. 

In this work we explore mesh-based GNN architectures; the latter exploit local (or short-range) interactions between neighboring nodes to estimate the response. When the nodes of a GNN are associated to a PDE computational domain, interactions between nodes can be interpreted as spatial interactions in the PDE domain. As most physical systems involve local interactions, GNN-based surrogates are among the best candidates to serve as effective simulator engines. However, when a physical system requires long-range interactions, such as in static solid mechanics problems, standard mesh-based, GNN approaches typically fail at capturing accurately the physics. This happens because in general the exchange of information between distant nodes requires a large number of message passing steps. This is unfavorable because of the poor scaling behaviors of GNNs with respect to the number of layers. In the context of graph-based neural operators, this problem has been addressed by using multi-resolution graphs to allow for faster propagation of information \cite{li2020multipole}. 

Here, we build on MeshgraphNets and propose two GNN architectures that overcome the challenge of long-range message passing without using a deep network. We introduce the Edge Augmented Graph Neural Network (EA-GNN) and the Multi-Graph Neural Network (M-GNN). In EA-GNN, we introduce ``virtual'' edges that yield faster information propagation resulting in better computational efficiency. In M-GNN we instead pursue a multi-resolution approach, inspired by the multi-grid method. Contrary to the work proposed in  \cite{multiscale} we obtain the low resolution graph by randomly removing nodes from the original mesh and by adding new edges to second or third order neighbors. In this process, the nodes keep their original coordinate attributes and the interactions with higher order neighbors are gradually added to the graph. Furthermore, in order to make GNNs generalizable to different geometries, we introduce an invariant simulation coordinate space by moving the physical systems to a simulation space that is invariant to translation and rotation (see \cite{liu2022ino} for a technique that enables the same properties in the context of graph-based neural operators). This process helps faster training of GNNs and allows for generalization to new geometries.

Our major contributions include:
\begin{itemize}
\item A novel coordinate transformation method to make the proposed GNN invariant to rotation and translation, and generalizable to new domain geometries.
\item An edge augmentation strategy that accelerates message passing across the graph, enabling more efficient training for problems involving long-range interactions.
\item A multi-graph approach, where information is passed through different resolution graphs in a hierarchical manner, resulting in faster propagation of messages across domain. 
\end{itemize}

\section{Background}
\label{litreview}

\subsection{Graph Neural Networks}
Graph Neural Networks (GNNs) are a class of deep learning methods that operate on graph structures consisting of nodes and edges \cite{zhou2020graph}.
GNNs  use message passing between neighboring nodes in the graph to model the interdependence of features at various nodes. They have been successfully used for prediction and classification tasks at the level of  graphs, edges, or nodes. In recent years, variants of GNNs such as graph convolutional network (GCNs) \cite{yao2019graph}, graph attention networks (GATs) \cite{velivckovic2017graph}, graph recurrent network (GRN) \cite{seo2018structured} have demonstrated ground-breaking performances on various tasks. The expressive nature and superior performance of GNNs have led to their application in a variety of domains where data is naturally represented with a graph structure, such as in particle physics \cite{shlomi2020graph}, high energy physics detectors \cite{ju2020graph}, power systems \cite{donon2019graph}, etc.  Additionally, GNN frameworks have been able  to simulate complex physical domains involving fluids, and rigid solids, and deformable materials interacting with one another \cite{pmlr-v119-sanchez-gonzalez20a}. In MeshGraphNets, the mesh physical domain is represented using a mesh, which is basically a graph on which GNNs learn to predict physics (see e.g., \cite{DBLP:journals/corr/abs-2010-03409}).

\subsection{Challenges in modeling deformation of Elastic and Hyper elastic Materials with GNNs}
\label{sec:mechanics}
In this work, we consider the stationary elasticity and hyper-elasticity problems, with a special focus on the Mooney-Rivlin hyper-elastic model \cite{BERGSTROM2015209}, even though our proposed approaches have the potential of performing well for a broad class of elliptic partial differential equations. We seek to train GNN models that capture the constitutive behavior of a physical system, rather than approximate the response for a fixed problem setting. To this end, we expect the proposed GNN surrogates to perform well on varying domain geometries, boundary conditions, loadings, and material properties. 
 
A challenging aspect of developing surrogates for stationary mechanics problems (and elliptic PDEs in general) is the fact that solution at any point depend on points that are farther away in the domain. In fact, the solution in these problems can be expressed as integrals over the entire domain. This means that the GNN must include connections between distant points within the network. In a way, the proposed architecture is  designed to enable fast message propagation between any points in the domain.

\subsection{MeshGraphNet for Physics Simulations}
\label{deepmind}
MeshGraphNets \cite{DBLP:journals/corr/abs-2010-03409} take advantage of the mesh representations, which is extensively used for finite element simulations for structural mechanics, aerodynamic problems, etc. With the development of adaptive meshing, accurate, high-resolution simulations can be carried out for deformation of complex geometries with highly irregular meshes. MeshGraphNets predict the dynamics of the physical systems by encoding the simulation state in meshes into a graph structure using an encoder, approximating the differential operators that underpin the internal dynamics of the physical systems using message passing steps of the graph neural network and using a decoder to extract the node level dynamics information to update the meshes. 

MeshGraphNet learns the forward model of the dynamic quantities of the mesh at time $t+1$ given the current mesh and meshes at previous time steps. The encoder encodes the current mesh into a graph, with mesh nodes becoming graph nodes and mesh edges becoming bi-directional edges in the graph. They also add extra edges, known as world edges, to learn external dynamics such as self-collision, contact etc which are non-local to the meshes. World edges are created by spatial proximity, within a fixed radius. Once the graph is created, node and edge attributes are generated and encoded using a learnable encoder network. Next, the processor, consisting of $L$ Graph Net (GN) blocks that can do message passing operations, update the world edges, mesh edges and the node features of the current graph. Finally, to predict the state at time $t+1$ from the state at time $t$, a decoder network, which is an MLP, is used to transform the updated node features to output features. 

MeshGraphNets can accurately and efficiently model the dynamics of the physical systems. It also has good generalization capability, and can be scaled up at inference time. However, they have not been tested on time-independent problems such as {static} solid mechanics problems.

In \cite{multiscale}, an improved version of MGNs, Multiscale MGN, was proposed with the purpose of recovering the concept of spatial convergence that characterizes mesh based discretizations and improving the training efficiency when refining the mesh. This work takes inspiration from Multigrid approaches from mesh based discretizations, but its intrinsic architecture and scope are different from the one presented in this paper.

\section{Methods}
\label{models}

As explained in the previous section, solving time-independent solid mechanics problems requires fast propagation of information across all the degrees of freedom, since the deformation, or stress, of every point in space depends on all the other points. Guaranteeing such propagation becomes computationally challenging as the resolution of the computational domain increases. To address this challenge, when using GNNs as numerical solvers, a natural solution is to increase the depth of the network. However, this approach is computationally very expensive and may result in over-smoothing \cite{zhao2019pairnorm}, making the model harder to train, or {causing} vanishing/exploding gradients \cite{wu2021representing}. To achieve fast propagation of information while avoiding these pitfalls, we propose two approaches: (a) Edge-Augmentated Graph Neural Network , and (b) Multi Graph Neural Network. In the following we describe each model separately.

\begin{figure}[t!]
    \begin{center}
 \includegraphics[width=0.95\columnwidth]{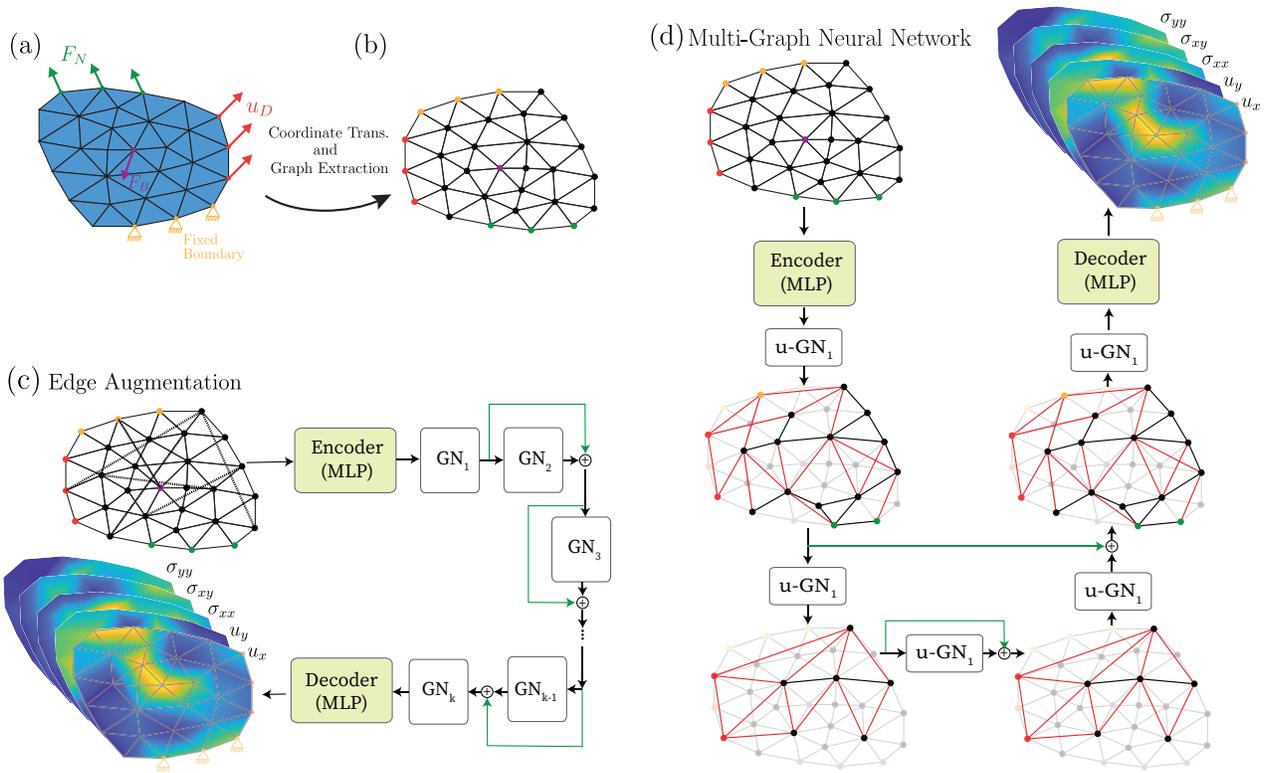}
    \caption{
    (a) Schematic of a 2D time-independent (static) solid mechanics problem. 
    (b) Physical domains are transferred to a normalized coordinate. 
    (c) Schematic of Edge-Augmented Graph Neural Networks (agumented edges are marked as dashed line) where the normalized graph is passed to an MLP Encoder and multiple passes of GN blocks and then a decoder to recover the displacement and stress vectors across the domain. 
    (d) Schematic of Multi-Graph Neural network, where features after encoding, pass through M-Net blocks with down-sampling/upsampling of the mesh. 
    }\vspace{-1mm}
    \label{fig:graphical-summary}
    \end{center}
\end{figure}

(a) \textbf{Edge Augmentation}: We refer to the first approach as  ``edge augmentation'' of the existing graph. As the graph is generated from a mesh, nodes are only connected to their neighbors; if two nodes are $n$ hops away, it requires $n$ GN blocks for the message to be passed between these nodes \cite{hamiltonbook2}. Thus, as $n$ increases, a deeper network is required to achieve broad message propagation. To avoid increasing the depth of the network, we increase the non-locality of the graph by introducing augmented edges between randomly selected nodes (see Fig. \ref{fig:graphical-summary}c). This edge augmentation significantly reduces the number of hops required for the nodes to gather messages from distant nodes, thus resulting in a smaller number of GN blocks for faster message propagation. 

(b) \textbf{Multi Graph}: This approach draws inspiration from multigrid methods for PDEs, whose foundation is a hierarchy of discretizations (or meshes) \cite{briggsbook}. By performing a series of graph down-sampling operations followed by up-sampling operations, we apply the multigrid concept to graphs and pass information across the computational domain in a hierarchical manner. In practice, the down-sampling operations correspond to mesh coarsening and result in smaller graphs connecting points that are far away in the domain. On the other hand, during up-sampling the information is redistributed across the entire graph. This approach resembles the graph U-Net architecture used in \cite{pmlr-v97-gao19a} for image processing tasks such as image classification and segmentation.

For both approaches, we assume that the computational domain is a triangular mesh which undergoes a transformation into a graph as detailed in Section \ref{data_gen} and illustrated in Figure \ref{fig:graphical-summary}.

\subsection{Edge Augmented Graph Neural Network (EA-GNN)}
\label{edge_aug_model}

With the purpose of achieving faster propagation of information between nodes we select a set of nodes from the graph and, if no edges exist between them, we add a bi-directional edge. We point out that, since this edge is not similar to other edges, we add an extra feature to the edge attributes to indicate that the edge belongs to the augmentation set. The number of additional edges in the augmented graph is a hyper parameter that is manually tuned during the training. Moreover, vertices of augmented edges are selected randomly, by following a uniform distribution.

The proposed EA-GNN has the same high-level network architecture as the MeshGraphNet  \cite{DBLP:journals/corr/abs-2010-03409} whose building blocks are: (i) an Encoder, (ii) $m$ GN blocks, and (iii) a Decoder. As described in Section \ref{data_gen}, for the time-independent problems considered in this work, with the purpose of making the graph translation and rotation invariant, before the graph is created, the mesh is {\it transformed} to be in the principal axis coordinate system (referred to as simulation coordinate system). Then, the mesh is {\it converted} into a graph structure by identifying the vertices of the mesh as nodes and the connections between the vertices in the mesh as the edges (see Fig. \ref{fig:graphical-summary}a,b).  The node attributes are nodal positions in simulation coordinates ($x$, $y$), nodal type (interior or boundary), type of the boundary condition applied (Dirichlet homogenous and non-homogenous, Neumann), direction and magnitude of the boundary conditions, a flag for body forces and its magnitude and direction all in simulation coordinate system. The edge attributes are the Euclidean distance between the nodes, and positional difference in $x$ and $y$ directions, i.e., $\Delta x = x_2 - x_1$ and $\Delta y = y_2 - y_1$.

Once the graph is generated, the node attributes and edge attributes are encoded into a latent space through an encoder. The encoded nodes and edges attributes are then passed through $m$ GN blocks and updated. Each GN block consists of an edge update module and a node update module. Let $\mathbf{u'}_i$ be the encoded node attribute vector of node $i$ and $\mathbf{u'}_j$ be the encoded node attribute vector for node $j$, such that $j \in M(i)$, where $M(i)$ denotes the neighborhood of node $i$. Let $\mathbf{e'}_{ij}$ be the encoded edge attribute vector for the edge between $i$ and $j$. The edge update module, $\chi$, is a MLP that receives the attributes of the nodes connecting the edge along with the edge attributes and returns the updated attributes, i.e. $\mathbf{e'}_{ij} = \chi(\mathbf{e'}_{ij})$. 

The node update module consists of three MLPs, $\phi$, $\gamma$ and $\beta$; the encoded node attributes of node $i$, $\mathbf{u'}_i$ are updated as follows:
\begin{equation}
    \begin{aligned}
    \label{eq:node_update}
    & \mathbf{m}_{ij} = \phi (\mathbf{u'}_i, \mathbf{u'}_j, \mathbf{e'}_{ij})\\
    & \mathbf{u'}_{i} = \gamma (\mathbf{u'}_i, \frac{1}{N}\sum_{j \in M(i)}\mathbf{m}_{ij})\\
    & \mathbf{u'}_{i} = \mathbf{u'}_{i} + \beta (\mathbf{u'}_{i}).
    \end{aligned}
\end{equation}
Between each GN block, a skip connection is added to avoid over-smoothing \cite{wu2021representing}.  The following equation describes the update for two consecutive GN blocks of the network, $GN_k$ and $GN_{k+1}$:
\begin{equation}
    \begin{aligned}
    \label{eq:GNupdate}
    &\mathbf{u'}^{(k)}, \mathbf{e'}^{(k)} = GN_{k}(\mathbf{u'}^{(k-1)},\mathbf{e'}^{(k-1)})\\
    &\mathbf{u'}^{(k)} = \mathbf{u'}^{(k)} + \mathbf{u'}^{(k-1)}\\
    &\mathbf{u'}^{(k+1)}, \mathbf{e'}^{(k+1)} = GN_{k+1}(\mathbf{u'}^{(k)},\mathbf{e'}^{(k)})\\
    &\mathbf{u'}^{(k+1)} = \mathbf{u'}^{(k+1)} + \mathbf{u'}^{(k)}.
    \end{aligned}
\end{equation}
Finally, the updated node attributes, $\mathbf{u'}$ is passed through a decoder to return the nodal deformation or stress values. Both the encoder and decoder functions are  MLP networks.

\subsection{Multi-Graph Neural Network (M-GNN)}

This approach takes inspiration from multigrid solvers and relies on hierarchical learning. The architecture has three components: (i) an Encoder, (ii) an {M-Net} block, and (iii) a Decoder (see Fig. \ref{fig:graphical-summary}d). Encoder and Decoder have the same network architecture and functionality as in EA-GNN; however, for reasons that are clarified later in this section, here we do not consider edge attributes.

The {M-Net} block starts with a single graph network block GN, to update the encoded node attributes received from the encoder. The GN block uses GraphSAGE operator \cite{NIPS2017_5dd9db5e} for the node update. The encoded node attributes, $\mathbf{u'}$, are updated through GN block as follows.
\begin{equation}
    \begin{aligned}
    \label{eq:sage_update}
    & \mathbf{u'}_{i} = \text{GraphSAGE} (\mathbf{u'}_i, \mathbf{u'}_{j}), \forall j \in M(i)\\
    & \mathbf{u'}_{i} = \mathbf{u'}_{i} + \beta (\mathbf{u'}_{i}),
    \end{aligned}
\end{equation}
where, $\beta$ is a MLP. The updated node attributes, $\mathbf{u'}$, are then passed through a series of alternating layers of down-sampling operations and GN blocks. This is followed by a series of alternating up-sampling operations and GN blocks. The number of down-sampling and up-sampling layers is determined by the multi-graph depth hyperparameter, $d$. The output of the down-sampling layer is added to the output of the corresponding up-sampling layer leading to the same mesh refinement (or graph size). Note that all the GN block updates have the same structure reported in Equation \ref{eq:sage_update} and that the final up-sampling layer is such that the graph recovers its original size.

The down-sampling layer down-samples the data by adaptively selecting a subset of nodes corresponding to a coarser mesh; the number of nodes that are down-sampled is determined by the hyperparameter $r$, indicating the down-sampling ratio. For the adaptive selection of nodes, we use the ``U-net sub-sampling'' algorithm \cite{pmlr-v97-gao19a}. Specifically, the node attributes are projected onto the trainable vector $\mathbf{p}$ using the scalar projection ${\bf u}_i^T{\bf p}$ and top $k$ nodes are selected based on the projected values. Since the scalar product measures the amount of information retained by node $i$ when projected onto $\mathbf{p}$, 
sampling the top $k$ nodes ensures that the smaller graph retains the maximum information. 
The up-sampling operation up-samples the graph by recording the locations of nodes selected in the corresponding down-sampling layer and uses this information to place the nodes back to their original positions in the graph.

In order to make sure that there are no disconnected nodes after down-sampling, as well as to improve the connectivity between the nodes, we compute the $l^{th}$ graph power, similarly to \cite{pmlr-v97-gao19a}, and use the resulting graph. This operation builds links between nodes which are at-most $l$ hops away in the graph. This could be done by multiplying the adjacency matrix of the graph by itself $l$ times. For the training, we choose $l=3$, and use the augmented graph with better connectivity for every down-sampling layer. This step is particularly important in mesh-based physics simulations for uninterrupted propagation of information between nodes.

The GraphSAGE operator \cite{NIPS2017_5dd9db5e}, used in the GN block, is a message aggregation algorithm which considers only the node attributes. In fact, as nodes are down-sampled, existing edges are lost and new edges are introduced to improve the connectivity. This makes updating edge attributes (for existing and newly established edges) a nontrivial task. Since the nodal positions are considered as node attributes, the model can calculate the edge attributes such as Euclidean distance and positional difference indirectly from the node attributes. Thus no information is lost due to the removal of edge attributes.

\section{Results and Discussion}
\subsection{Data Generation}
\label{data_gen}

The data generation process consists of three steps: (i) mesh generation, (ii) finite element simulations (iii) graph transformation. We generate two-dimensional random geometries using Bezier curves to ensure variability and nonlinearity in the domains. Different boundary conditions (Dirichlet and Neumann)  are assigned at randomly selected locations on the boundary. The length, location, magnitude and direction of the boundary conditions are randomly selected for each geometry. We also assign a body force at a randomly selected interior location of the domain. The mesh is generated using the \texttt{Gmsh} package in python \cite{geuzaine2009gmsh}. Using Abaqus \cite{manual2020abaqus}, we carry out finite element simulations to obtain nodal deformation and stress values. 


\paragraph{Coordinate Transformation}
As the output values (deformation and stress) are coordinate dependent and every sample in the training set is characterized by a different geometry, it is necessary to assign the nodal coordinates as node attributes. However, this makes the graph translation and rotation variant, i.e. if the mesh is translated or rotated, the network would consider it as a different geometry and outputs different deformation and stress values. This makes the training process difficult as it requires redundant data in training to make the model invariant to rotation and translation.To resolve this issue, we use group equivariance as our inductive bias where we ensure the graph is invariant to translation and rotation by transforming the geometry into the principal axis coordinate system. As such, the nodal coordinates stay the same when the geometry is rotated or translated, and as a result in the transformed domain is invariant to the rotation and translation. From now on, we refer to these coordinates as ``simulation coordinates'' (SC).

\paragraph{Data Augmentation}
To improve the generalization capability of the model with respect to unseen geometries, we introduce two data-augmentation strategies. First, we use both the Delaunay and the ``Packing of Parallelograms'' algorithms for mesh generation. Second, we add noise to nodal coordinates by shifting the nodal coordinates by adding to each node a random value sampled from the normal distribution $\mathcal{N}(0,0.01)$ which is equivalent to 10\% of edge distance on average.

\paragraph{Data Conversion} 
Nodal deformations and stress values are assigned to the node and edge attributes for the mesh-to-graph conversion outlined in section \S\ref{edge_aug_model}. For GNN with edge augmentation, augmented edges are added to the existing edge connections together with an augmentation flag ($1$ for augmented edges and $0$ otherwise) as edge attribute. The node and edge attributes along with the edge connections are used to generate the graph objects using PyTorch Geometric.
 
\subsection{Model Architectures}

\subsubsection{EA-GNN}
\label{edge_aug_arch}

During graph creation, edge augmentation is done by sampling nodes from the graph and adding bi-directional edges between them, if there are no existing edges. The percentage of augmented edges is determined by the hyper parameter augmented-edge-percentage, $A_{perc}$ . We use $A_{perc} = 20\%$ for our experiments. 

The encoders for the node and edge attributes of the augmented graph are MLPs with a single hidden layer and a ReLU activation function. The MLP for node attributes has a network architecture $14-64-128$ (i.e., 14 input features, 64 nodes in the hidden layer and 128 output features) and that for edge attributes has network architecture $4-64-128$. The encoded graph is then passed through a series of alternating layers of Graph Net (GN) blocks and MLP layers. A total of 6 GN blocks are added in the network. Each GN block consists of  (a) \textit{Edge update module} - The edge update function, $\chi$, is a MLP, with a single hidden layer architecture $3\times128-128-128$, and ReLU activation function. The size of the input is due to concatenating the edge attributes ($128$ features) and the attributes of the two nodes connected by the edge ($2\times128$ features). The output consists of the updated edge attributes of size $n_e \times 128$, where $n_e$ is the total number of edges.  (b) \textit{Node update module} - This module consists of three functions, $\phi$ for message passing, $\gamma$ for message aggregation and node update and $\beta$ for the final node update. All the three functions are MLPs with a single hidden layer and ReLU activation functions.  The function $\phi$ has network architecture  $2\times128-128-128$, where the input features are the concatenation of the attributes of the neighboring node and the corresponding edge attributes. The messages from all the neighboring nodes are added and concatenated to the node attributes of the selected node and the result is given as the input to $\gamma$ to calculate the node attributes of the corresponding nodes. Thus, the network architecture of $\gamma$ is  $2\times128-128-128$. The output from $\gamma$ is passed to $\beta$ to get the updated node attributes as shown in Equation \ref{eq:node_update}. The MLP representing $\beta$ has network architecture $128-128-128$.

The parameters of all the four functions ($\chi$, $\phi$, $\gamma$, $\beta$) are shared across all the GN blocks.  After each GN block a skip connection is added as shown in Equation \eqref{eq:GNupdate}. The updated node attribute from the final GN block is the input to the decoder, a MLP with a single hidden layer, ReLU activation functions, and architecture $128-64-2$ for the displacement ($u_x$ and $u_y$) and $128-64-3$ for the stress ($\sigma_{xx}$, $\sigma_{xy}$ and $\sigma_{yy}$). We use a dropout layer after the encoder and between each GN block with a dropout percentage of $0.1$ to reduce overfitting.

\subsubsection{M-GNN}
M-GNN has an encoder architecture similar to that of EA-GNN. However, since edge attributes are not used, the encoder is required only for the node attributes. The encoder is a MLP with architecture $14-64-128$. The encoded graph is passed through a message aggregation block followed by a ReLU activation layer. The message aggregation block used here is the GraphSAGE operator \cite{NIPS2017_5dd9db5e}, which updates the node attributes for the full graph.

This is followed by a series of {down-sampling modules consisting} of the following operations: (a) \textit{Connectivity enhancement}-- The connectivity of the graph is enhanced by connecting the nodes that are $3$ hops away. This is done by multiplying the adjacency matrix, $A$ by itself twice, i.e. $A_{upd} = A \times A \times A$.  (b) \textit{Down-sampling}-- A layer that projects the node attributes onto a one-dimensional, trainable projection vector $\mathbf{p}$, and samples $k$ nodes based on the projected values. The value of $k$ is determined by the down-sampling ratio, $r$. For the experiments, we have set $r=0.6$. (c) \textit{Node update}-- A block consisting of the GraphSAGE operator and the function $\beta$, as described in Equation \ref{eq:sage_update}. 

The number of down-sampling modules is determined by the multi-graph depth hyperparameter, $d$. After tuning of the hyperparameters, we have set this value to be $d=3$. These are followed by same number of up-sampling modules, consisting of the following operations: (a) \textit{Skip Connection}-- The node attributes from the corresponding down-sampling layer are added to the updated node attributes. (b) \textit{Up-sampling}-- A layer that adds the previously removed nodes back to the graph, with their updated node attributes. (c) \textit{Node update}-- A block consisting of the GraphSAGE operator and the function $\beta$, as described in Equation \ref{eq:sage_update}. 

The final pooling module restores the full graph with all the nodes with updated node attributes. This is decoded to return the nodal displacement or stress using the Decoder, a network with the same architecture used for EA-GNN. Similar to EA-GNN, we use a dropout layer after the encoder and between each GN block with a dropout percentage of $0.1$ to reduce overfitting.

\subsection{Experimental Set up}
\label{experiments}
By using the data generation process detailed in section \S\ref{data_gen}, we generate the following three datasets and carry out data augmentation as explained in Section \S\ref{data_gen}. (a) We generate 3000 random geometries and run 10 variations of boundary conditions per geometry assuming a homogeneous and linear elastic material with constant Young's modulus $E=100.0$ and Poisson's ratio $\nu=0.3$. This dataset is used to evaluate the ability of the model to capture geometrical nonlinearities. (b) We generate 5000 random geometries and assume a nonlinear hyper-elastic material. We again consider 10 different boundary conditions per geometry. The nonlinear material model of choice is the Mooney-Rivlin model \cite{liu2012note}, with material properties $C_{01}=0.3$ and $C_{10}=0.03$ \cite{kumar2016hyperelastic}. Training with this dataset enables us to test the network performance for system nonlinearities. (c)  We generate 10,000 random geometries with varying hyper-elastic material properties and again consider 10 different boundary conditions per geometry. The material properties for each geometry are randomly selected from a uniform distribution, $C_{01} \sim \mathcal{U}(0.034, 0.34)$ and $C_{10} \sim \mathcal{U}(0.005, 0.05)$. We use this dataset to train and test our architecture on generalization with respect to unseen materials.

For all the experiments, 70\%, 10\%, 20\% of the data is used for training, validation, and testing respectively. For each dataset we train four networks to predict the nodal deformations, $u_x$ and $u_y$. As a baseline model (B) we consider MeshGraphNet. Since our simulation coordinate transformation plays an important role in the generalization properties of our model, we further consider a version of MeshGraphNet trained after such a transformation and refer to it as B + SC. We additionally provide results for our Edge-Augmented Graph Neural Network using simulation coordinates (EA-GNN + SC), and Multi-Graph Neural Networks using simulation coordinates (M-GNN + SC). Similarly, the four networks are trained to predict nodal stress values, $\sigma_{xx}$, $\sigma_{xy}$ and $\sigma_{yy}$. 

While the baseline model is trained using mean squared error loss function as used in  \cite{DBLP:journals/corr/abs-2010-03409}, we train the networks B+SC, EA-GNN+SC and M-GNN+SC with a scaled mean absolute error loss function $L$. The scaling depends on a combination of the boundary conditions associated with each sample; formally, 
\begin{equation}
    \begin{aligned}
    \label{eq:loss}
    L(\theta)= \dfrac{1}{n}\sum_{n=1}^N (\|{\bf d}_n\|_{\ell^1}+\|{\bf n}_n\|_{\ell^1}) \|{\bf y}_n - \widehat{\bf y}_n\|_{\ell^1},
    \end{aligned}
\end{equation}
where $\bf y$ and $\hat{\bf y}$ are respectively the ground-truth and the predicted outputs and ${\bf d}_n$ and ${\bf n}_n$ are the Dirichlet loading and Neumann displacement vectors of the $n$-th sample.

For all networks we use the Adam optimizer and we train all the networks for 1500 epochs on a Tesla V100-SXM2. For B, B+SC and EA-GNN+SC, we prescribe a learning rate ranging from 1E-4 to 1.5E-4 and a weight decay of 1E-5. The learning rate is decreased using the cosine annealing scheduler with warm restart \cite{loshchilov2016sgdr}, which avoids stagnation in local minima. For M-GNN+SC, we use a learning rate ranging from 2E-3 to 3E-3 with a weight decay of 1E-6.

\subsection{Numerical Results}

We test our proposed models on three datasets detailed in section \S\ref{experiments} and compare them with the baseline models. In order to compare the prediction power of our models, we define relative error as
$$
e(f) = \dfrac{\|\widehat f - f\|_{\ell^1}}{\|f\|_{\ell^1}}.
$$
{where $f$ is the variable of interest (either displacement values $u_x,u_y$ or stress values $\sigma_{xx}, \sigma_{yy}, \sigma_{xy}$) and $\widehat{f}$ is the predicted value of $f$ using our network. 

%

The relative errors on the test dataset for the prediction of nodal displacement values ($u_x$ and $u_y$) and nodal stress values ($\sigma_{xx}$, $\sigma_{yy}$ and $\sigma_{xy}$) are presented in Tables \ref{table:1} and \ref{table:2}. Here, E stands for linear elastic model and HE for hyper-elastic (Mooney-Rivilin) material model. Additionally, S stands for single material and it means that a single material is used across the entire dataset, whereas V stands for varying material and it means that varying materials are used across the data set.}

\vspace{4mm}
\begin{minipage}[H]{0.45\textwidth}
\centering
\begin{center}
\begin{small}
\begin{sc}
\begin{tabular}{lccccr}
\toprule
Mat. & Model & Param. & $e(u_x) \downarrow$ & $e(u_y) \downarrow$  \\
\midrule
E,S & B & 7.2e5 & 0.64 & 0.63\\
E,S & B+SC & 7.2e5 & 0.25 & 0.26\\
E,S & EA-GNN+SC & 7.2e5 & \textbf{0.05} & \textbf{0.05}\\
E,S & M-GNN+SC  & 2.8e5 & 0.13 & 0.13\\
\midrule
HE,S & B & 7.2e5 & 0.81 & 0.81 \\
HE,S & B+SC & 7.2e5 & 0.33 & 0.34 \\
HE,S & EA-GNN+SC & 7.2e5 & \textbf{0.08} & \textbf{0.09}\\
HE,S & M-GNN+SC & 2.8e5 & 0.16 & 0.16  \\
\midrule
HE,V & B & 7.2e5 & 0.73 & 0.74\\
HE,V & B+SC & 7.2e5 & 0.28 & 0.29\\
HE,V & EA-GNN+SC & 7.2e5 & \textbf{0.09} & \textbf{0.09} \\
HE,V & M-GNN+SC & 2.8e5 & 0.16 & 0.15 \\
\bottomrule
\end{tabular}
\captionof{table}{Relative error in predicting nodal displacements $u_x$ and $u_y$ given different linear and nonlinear material selection and models.}
\label{table:1}
\end{sc}
\end{small}
\end{center}
\end{minipage}
\begin{minipage}[H]{0.45\textwidth}
\centering
\begin{center}
\begin{small}
\begin{sc}
\begin{tabular}{lccccr}
\toprule
Mat. & Model  & $e(\sigma_{xx}) \downarrow$ & $e(\sigma_{yy}) \downarrow$  & $e(\sigma_{xy})\downarrow $ \\
\midrule
E,S & B  & 0.78 & 0.79 & 0.49\\
E,S & B+SC  & 0.30 & 0.30 & 0.19\\
E,S & EA-GNN+SC  & \textbf{0.09} & \textbf{0.09} & \textbf{0.05}\\
E,S & M-GNN+SC  & 0.17 & 0.18 & 0.11\\
\midrule
HE,S & B  & 0.88 &0.88 & 0.64\\
HE,S & B+SC  & 0.30 &0.32 & 0.29\\
HE,S & EA-GNN+SC  & \textbf{0.11} & \textbf{0.13} & \textbf{0.09}\\
HE,S & M-GNN+SC  & 0.18 & 0.19 & 0.15 \\
\midrule
HE,V & B  & 0.92 & 0.91 & 0.85\\
HE,V & B+SC  & 0.39 & 0.39 & 0.29\\
HE,V & EA-GNN+SC  & \textbf{0.12} & \textbf{0.12} & \textbf{0.09}\\
HE,V & M-GNN+SC  & 0.18 & 0.20 & 0.16\\
\bottomrule
\end{tabular}
\captionof{table}{Relative error in predicting nodal stress $\sigma_{xx}, \sigma_{xy}, \sigma_{yy}$ given different linear and nonlinear material, and neural network models.}
\label{table:2}
\end{sc}
\end{small}
\end{center}
\end{minipage}
\vspace{3mm}

As shown in Tables \ref{table:1}, \ref{table:2}, both EA-GNN and M-GNN perform significantly better than the baseline models either with or without coordinate transformation, i.e., B and B+SC respectively. This shows that both the edge augmentation and the multi-graph modifications have the ability to further improve message passing. We further observe that by adding the coordinate transformation, the performance of the baseline model improves significantly, however, it is still very large compared to our proposed approaches. This comparison allows us to distinguish the improvement brought by the coordinate transformation and the graph modifications. It is interesting to see that the edge augmentation technique consistently performs better than all other methods, as highlighted by the bold values used for the best performing models. Comparing the rows of single linear elastic material (E,S) with  single nonlinear hyper-elastic materials (HE,S), we find that the performance of all the models slightly decrease and the relative absolute errors increases as expected. The non-linearity introduces new complexities into the system which requires more data for the learning. Note that adding further complexity by varying the nonlinear material among test cases (i.e., HE, V), we find that the performance stays the same and the proposed model can generalize to unseen material properties. The similar test performance of our models on HE,V test-dataset, as opposed to the performance of the baseline methods, shows the generalization capabilities of our model, which is indeed able to learn the true physics and predict accurate results for unseen nonlinear material properties.

\begin{figure}[ht]
\vskip 0.2in
\begin{center}
\centerline{\includegraphics[width=0.95\columnwidth]{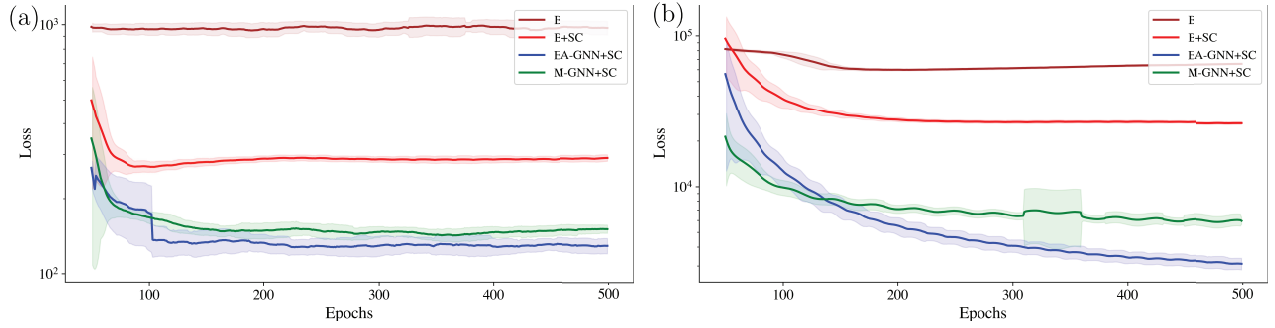}} 
\caption{Loss on testing data for predicting (a) deformation and (b) stress  for elastic model with single material for four models - B, B+SC, EA-GNN+SC and M-GNN+SC}
\label{fig_loss_deformation_stress}
\end{center}
\vskip -0.2in
\end{figure}

In addition to the relative error values, we show the loss values for the test dataset in correspondence of all the models for both the displacement $u_x, u_y$ and the stress values $\sigma_{xx}, \sigma_{yy}, \sigma_{xy}$ in Figs. \ref{fig_loss_deformation_stress}a and \ref{fig_loss_deformation_stress}b respectively. The lines represent the moving average, while the shadow corresponds to the variance. The sudden jumps in the loss values observed in M-GNN and EA-GNN correspond to the annealing process. Overall, we find that the coordinate transformation, improves the training of the baseline model, however, great boost in performance is only achieved when considering Edge-Augmentation or Multi-Graph approach. 



Next, in order to evaluate the generalization capability of our models and confirm that the models have indeed learned the true physics, we test the trained networks on out-of-distribution datasets, i.e. datasets that are not from the distribution used in training. To this end, we consider the three out-of-sample distributions:  (a) Smaller/larger physical domains characterized by half or double the domain size with the same mesh characteristic length. The results for scale = 0.5 and 2.0 are shown in Table \ref{table_u_pred_oos}. (b) New Dirichlet and Neumann boundary conditions applied at different sections of the boundary. Specifically, Dirichlet and Neumann boundary conditions are applied at disconnected sets of boundary nodes as opposed to connected sets. The results for  out-of-distribution boundary conditions are shown Table \ref{table_u_pred_oos} with the label BC.  (c) A combination of all the above along with random rotation and translation, i.e. the graphs from the above data are translated and rotated to a different coordinate system to evaluate if the models are rotation and translation invariant. The results for rotation and translation are also presented in Table \ref{table_u_pred_oos} with label Rot+Tra.

\begin{table}[H]
\vskip 0.15in
\begin{center}
\begin{small}
\begin{sc}
\begin{tabular}{lcccr}
\toprule
Model  &  B & B+SC & EA-GNN+SC & M-GNN+SC \\ 
\midrule
Scale = 0.5\\
\midrule
$e(u_x)$ & 0.87 $\pm$ 0.56 & 0.47 $\pm$ 0.29 & \textbf{0.09 $\pm$ 0.07} & 0.20 $\pm$ 0.11 \\
$e(u_y)$ & 0.96 $\pm$ 0.64 & 0.47 $\pm$ 0.33 & \textbf{0.09 $\pm$ 0.07} & 0.20 $\pm$ 0.11\\
\midrule
Scale = 2\\
\midrule
$e(u_x)$ & 1.09 $\pm$ 0.67 & 0.85 $\pm$ 0.41 & \textbf{0.22 $\pm$ 0.12} & 0.29 $\pm$ 0.11 \\
$e(u_y)$ & 0.95 $\pm$ 0.51 &  0.76  $\pm$ 0.33 & \textbf{0.21 $\pm$ 0.10} & 0.28 $\pm$ 0.09\\
\midrule
BC\\
\midrule
$e(u_x)$ & 0.87 $\pm$ 0.55 & 0.57 $\pm$ 0.42 & \textbf{0.11 $\pm$ 0.08} & 0.22 $\pm$ 0.10 \\
$e(u_y)$ & 0.95 $\pm$ 0.65 &  0.60  $\pm$ 0.43 & \textbf{0.13 $\pm$ 0.09} & 0.24 $\pm$ 0.10\\
\midrule
Rot + Tra\\
\midrule
$e(u_x)$ & 0.91 $\pm$ 0.69 & 0.57 $\pm$ 0.42 & \textbf{0.11 $\pm$ 0.08} & 0.22 $\pm$ 0.10 \\
$e(u_y)$ & 1.04 $\pm$ 0.78 &  0.60  $\pm$ 0.43 & \textbf{0.13 $\pm$ 0.09} & 0.24 $\pm$ 0.10\\
\bottomrule
\end{tabular}
\caption{Relative error in predicting nodal deformation for various out of sample distributions of data}
\label{table_u_pred_oos}
\end{sc}
\end{small}
\end{center}
\vskip -0.1in
\end{table}

The first observation is that the training based on simulation coordinates helps the generalization on all fronts. Next, we observe that EA-GNN consistently outperforms other architectures, followed by M-GNN, both evaluated in simulation coordinate system. Both EA-GNN and M-GNN have relative errors close to the in-distribution testing errors from Table \ref{table:1} for half scaling. The models perform worse for the double scaling case, as almost half of the graph corresponds to coordinate positions the model has not seen before. The errors for the rotation and translation do not change for B+SC, EA-GNN+SC and M-GNN+SC as the graphs are transformed to the simulation coordinate system before they are fed into the model and as expected the models perform well, as long as they are trained using simulation coordinates. Across all the cases, the baseline model B consistently performs the worst, with errors 1.5 to 2 times bigger than B+SC. This is the only model which is not invariant to translation and rotation, as can be observed from the higher errors for rotation and translation case for B.

From these experiments, we can infer that both EA-GNN and M-GNN are able to successfully learn the underlying physics of the data and generalize well to unseen domains, geometries, and boundary conditions, which the baseline model fails at. Training and evaluating the models in simulation coordinate system enables them to be invariant to rotation and translation, which makes these models an effective tool for faster and accurate physics simulations specially for time-independent physics simulations that require long-range interaction between different parts of the domain.

\section{Conclusion}

State-of-the-art GNN architectures are incapable of efficiently capturing static mechanics behavior as they cannot efficiently describe long-range interactions. In this paper we show that Edge augmented GNNs and Multi-GNNs can capture accurately and efficiently the constitutive behaviour of static elastic and hyper elastic materials thanks to their enhanced connectivity. Furthermore, by learning the physics in a reference coordinate system, our models are automatically rotation and translation invariant. With several numerical tests, we show that both our proposed architectures learn time-independent solid mechanics efficiently and generalize well to unseen materials, boundary conditions and domains. We note that the proposed approach for coordinate transformation can be easily used in any graph-based architecture, as we demostrate for MeshGraphNets. This is the first time graph based neural network surrogates have been successfully used for solving time-independent solid mechanics problems in a translation and rotation invariant manner and with generalization guarantees. Our approach represents an easily implementable solution for learning challenging time-independent physical systems using deep learning and, as such, it is a good candidate for simulating complex static systems in science and engineering.

\bibliography{ScientificReport}



\section{Acknowledgements}


R.G. and A.Z. would like to thank Prajjwal Jamdagni for the great and detailed discussions on FEM numerical simulations. A.Z and V.S  extend their gratitude to Abhishek Sharma who provided encouragement and support in the initiation of this project, and  whose guidance and mentorship will always be appreciated.

\section*{Author contributions statement}

A.Z. and V.S. initiated the concept for this study. R.G. and H.R. were responsible for conducting the finite element method simulations necessary for training the AI models. R.G., M.D., and A.Z. devised the AI architectures and R.G. implemented and carried out the AI model training and testing. All authors contributed to analysis of the results, writing of the manuscript, and reviewing the final version.

\section*{Additional information}

To include, in this order: \textbf{Accession codes} (where applicable); \textbf{Competing interests} (mandatory statement). 

The corresponding author is responsible for submitting a \href{http://www.nature.com/srep/policies/index.html#competing}{competing interests statement} on behalf of all authors of the paper. This statement must be included in the submitted article file.

\end{document}